\documentclass[a4paper,twoside]{article}

\usepackage{epsfig}
\usepackage{subcaption}
\usepackage{xcolor}
\usepackage{calc}
\usepackage{amssymb}
\usepackage{booktabs}
\usepackage{amstext}
\usepackage{amsmath}
\usepackage{multirow}
\usepackage{amsthm}
\usepackage{multicol}
\usepackage{pslatex}
\usepackage{apalike}
\usepackage[linesnumbered,ruled,vlined]{algorithm2e}
\usepackage{hyperref}
\usepackage{mathrsfs}
\usepackage[bottom]{footmisc}
\usepackage{SCITEPRESS}     % Please add other packages that you may need BEFORE the SCITEPRESS.sty package.

\hypersetup{
    colorlinks=true,
    citecolor=green
}

\begin{document}

\title{Enhanced Generative Data Augmentation for Semantic Segmentation \\
via Stronger Guidance}

\author{\authorname{Quang-Huy Che\sup{1,}\sup{2}, Duc-Tri Le\sup{1,}\sup{2}, Bich-Nga Pham\sup{1,}\sup{2},\\
Duc-Khai Lam\sup{1,}\sup{2} and Vinh-Tiep Nguyen\sup{1,}\sup{2,}\thanks{Corresponding author}}
\affiliation{\sup{1}University of Information Technology, Ho Chi Minh City, Vietnam}
\affiliation{\sup{2}Vietnam National University, Ho Chi Minh City, Vietnam}
\email{huycq@uit.edu.vn, 21522703@gm.uit.edu.vn, \{ngaptb, khaild, tiepnv\}@uit.edu.vn}
}


\abstract{Data augmentation is crucial for pixel-wise annotation tasks like semantic segmentation, where labeling requires significant effort and intensive labor. Traditional methods, involving simple transformations such as rotations and flips, create new images but often lack diversity along key semantic dimensions and fail to alter high-level semantic properties. To address this issue, generative models have emerged as an effective solution for augmenting data by generating synthetic images. Controllable Generative models offer data augmentation methods for semantic segmentation tasks by using prompts and visual references from the original image. However, these models face challenges in generating synthetic images that accurately reflect the content and structure of the original image due to difficulties in creating effective prompts and visual references. In this work, we introduce an effective data augmentation pipeline for semantic segmentation using Controllable Diffusion model. Our proposed method includes efficient prompt generation using \textit{Class-Prompt Appending} and \textit{Visual Prior Blending} to enhance attention to labeled classes in real images, allowing the pipeline to generate a precise number of augmented images while preserving the structure of segmentation-labeled classes. In addition, we implement a \textit{class balancing algorithm} to ensure a balanced training dataset when merging the synthetic and original images. Evaluation on PASCAL VOC datasets, our pipeline demonstrates its effectiveness in generating high-quality synthetic images for semantic segmentation. Our code is available at \href{https://github.com/chequanghuy/Enhanced-Generative-Data-Augmentation-for-Semantic-Segmentation-via-Stronger-Guidance}{this https URL}.}

\onecolumn \maketitle \normalsize \setcounter{footnote}{0} \vfill

\section{\uppercase{Introduction}}
\label{sec:introduction}

\begin{figure*}
    \includegraphics[width=\textwidth]{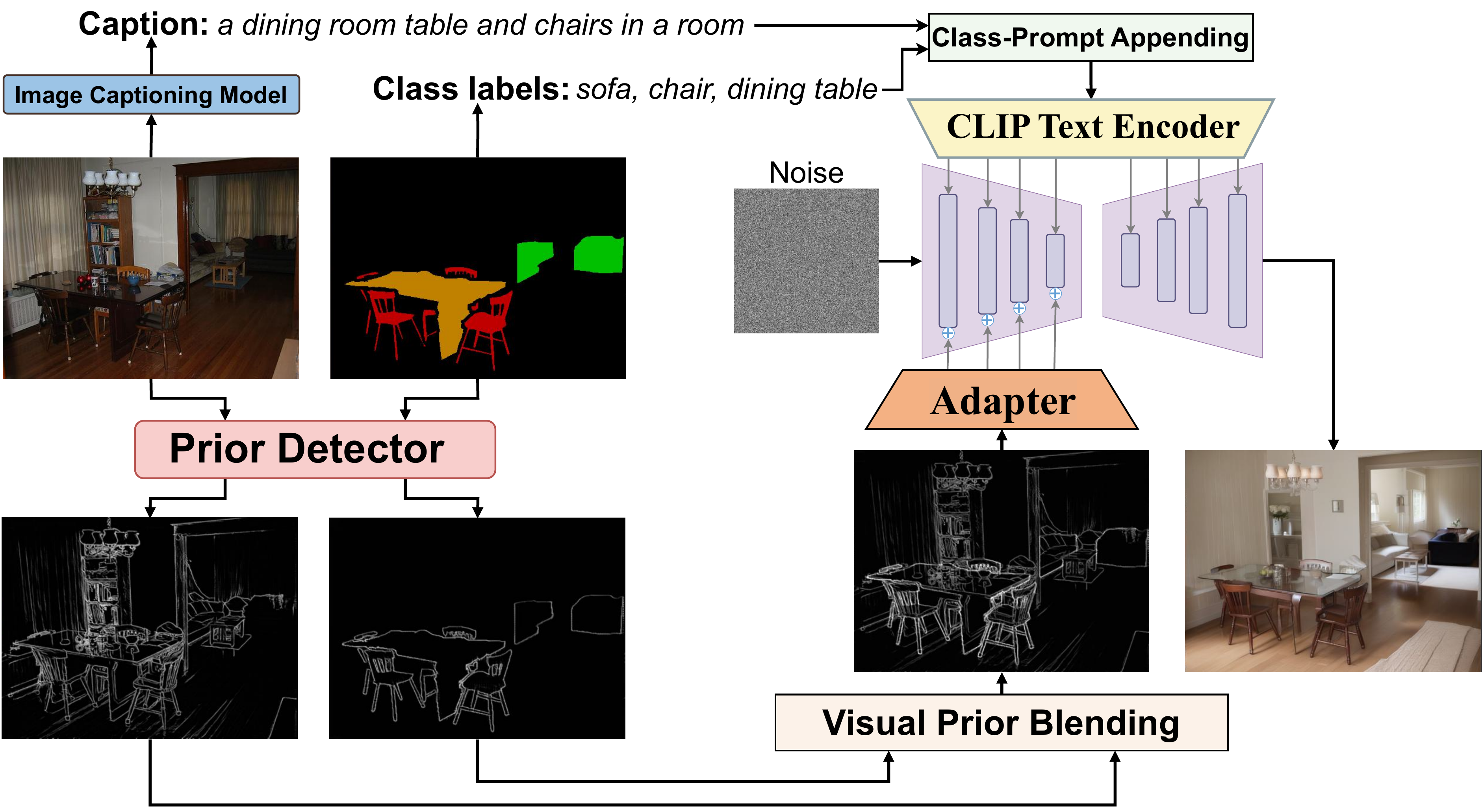}
    \caption{The controllable data augmentation pipeline for the semantic segmentation task combines our proposed methods: Class-Prompt Appending and Visual Prior Blending}
    \label{main}
\end{figure*}

Semantic segmentation is a fundamental computer vision task that involves classifying each pixel in an image. Deep learning models have significantly advanced semantic segmentation methods. These models are usually trained on large-scale datasets with dense annotations, such as PASCAL VOC \cite{voc}, MS COCO \cite{coco}, BDD100K \cite{bdd}, and ADE20K \cite{ade20k}. It is often necessary to re-label the data to address a specific task. However, labeling a new dataset to enable accurate model learning is time-consuming and costly, particularly for semantic segmentation tasks that require pixel-level labeling.

An alternative to enhancing data diversity without annotating a new dataset is data augmentation, which creates more training examples by leveraging an existing dataset. Commonly used data augmentation methods in semantic segmentation include rotating, scaling, flipping, and other manipulations of individual images. These techniques encourage the model to learn more invariant features, thereby improving the robustness of the trained model. However, basic transformations do not produce novel structural elements, textures, or changes in perspective. Consequently, more advanced data augmentation methods utilize generative models for different tasks \cite{aug_cls1,aug_cls2,aug_cls3,aug4,aug_wsss}. Generative models leverage the ability to create new images based on inputs such as text, semantic maps, and image guidance to specific tasks and data augmentation needs. Notably, Stable Diffusion (SD) models \cite{SD,SDXL} propose a method for conditional image generation that fusions textual information, bounding boxes, or segmentation masks to generate or inpaint images. Controllable Models \cite{controlnet,t2iadapter} further enhance the guided image generation capability by utilizing visual priors such as edges, depth, segmentation mask, human poses, etc.

The segmentation mask annotation can be easily computed in data augmentation using simple transformations (e.g. translation, scaling, flip). However, with generative models, data augmentation is more problematic as it requires generating new images while still matching the ground truth annotations. A straightforward approach to this challenge is to utilize the Inpainting model \cite{SD} to change the labeled regions in the images while keeping the rest of the remaining information. Although this method can enhance the diversity of labeled data, it does not ensure that the newly generated object matches the original structure, and the surrounding elements may lack diversity since they are left unchanged. Additionally, \cite{t2iadapter,controlnet,seggen1} propose image generation models that can be controlled via segmentation masks, making these methods highly suitable for efficient data augmentation in semantic segmentation tasks. However, generative models require training on semantic segmentation datasets, which limits their capacity to provide information beyond the scope of the training dataset.

% With a deep understanding of pre-trained generative models, we can identify their strengths with a vast knowledge base, such as generality and good controllability. In this paper, we propose using Controllable Generative models with prior visual without training on semantic segmentation datasets to augment data for semantic segmentation. The proposed methods ensure that the generated images match the original images in terms of the number of labels and their structures but with transformations in color, context, and style. To achieve high performance in data augmentation, we propose \textit{Visual Prior Blending} and \textit{Class-Prompt Appending} to enhance the visual representation of labeled classes. This combination method is depicted in Figure \ref{main}. In addition, we also use a \textit{class balancing algorithm} to control the number of classes during image generation so that when combining the synthetic dataset and the original data, the classes in the extended dataset are more balanced.

To address the issue of data augmentation for semantic segmentation using generative models, we can leverage a deep understanding of generative models to identify their strengths, such as their broad knowledge base, generalization capabilities, and structural control. In this paper, we propose using Controllable Generative models with prior visual without training on semantic segmentation datasets to augment data for semantic segmentation. \textbf{However, utilizing Controllable Generative models directly may present challenges, such as generated objects not strictly adhering to their original structure or a lack of labeled classes in the generated images. So, how can we overcome these limitations?} In this work, we propose \textit{Visual Prior Blending} to enhance the visual representation of labeled classes and \textit{Class-Prompt Appending} to construct text prompts that generate images containing all labeled classes. The combination of these two proposed methods, as shown in Figure \ref{main}, enables more effective data augmentation when utilizing Controllable Generative models, resulting in improved performance.

In this work, we analyze and propose methods to address these challenges when generating images using controllable generative models. 

Our main contributions are:

\begin{itemize}
    \item \textit{Class-Prompt Appending}: By combining “\textit{generated caption}” and “\textit{labeled classes}”, we use this method to construct text prompts containing comprehensive information about the image and its classes. This ensures that all labeled classes are fully represented in the synthesized image.

    \item \textit{Visual Prior Blending}: This method blends the visual priors from the original image and the segmentation masks. It aims to balance the information between the original image and the labeled objects, ensuring that the generated objects maintain the structure defined by the segmentation labels.

    \item In addition, we also use a \textit{class balancing algorithm} to control the number of classes during image generation so that when combining the synthetic dataset and the original data, the classes in the extended dataset are more balanced.

    \item We evaluate our proposed method on the VOC7/VOC12 datasets with various settings. The results demonstrate the effectiveness of the proposed approach, particularly in scenarios with limited data samples.
    
\end{itemize}

\section{\uppercase{Related Work}}

Data augmentation using generative models is commonly applied in classification tasks \cite{aug_cls1,aug_cls2,aug_cls3}. Unlike classification tasks, to augment data for Segmentation or Object Detection tasks, where the synthetic images must ensure the location of the objects. Inpainting model \cite{SD,SDXL,inpaint} is considered an option because it allows to specify what to edit as a mask. However, using an off-the-shelf inpainting model for data augmentation in semantic segmentation tasks may lead to shortcomings. There is no guarantee that the newly created objects align with the ground truth annotations, and the model may also tend to replace smaller objects with background elements.

\begin{figure*}[!t]
  \centering
    \includegraphics[width=\textwidth]{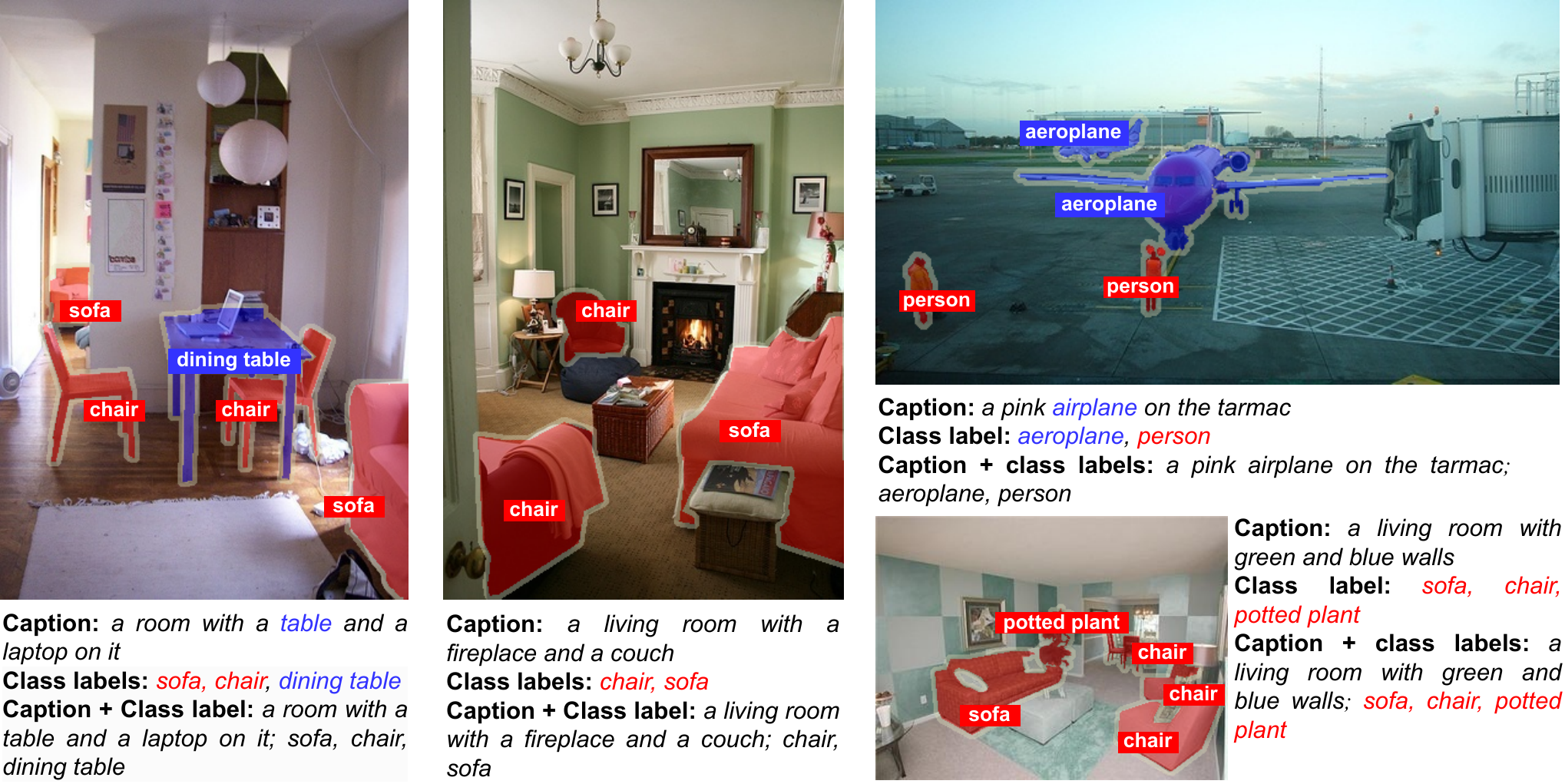}
    \caption{\textbf{Classes missing in the generated prompts:} {\color[HTML]{FE0000}Red} describes the missing classes in the generated prompt while {\color[HTML]{3531FF}blue} marks the ones appearing in the sentence.} 
    \label{loseclass}
\end{figure*}

\begin{figure*}[!t]
  \centering
    \includegraphics[width=\textwidth]{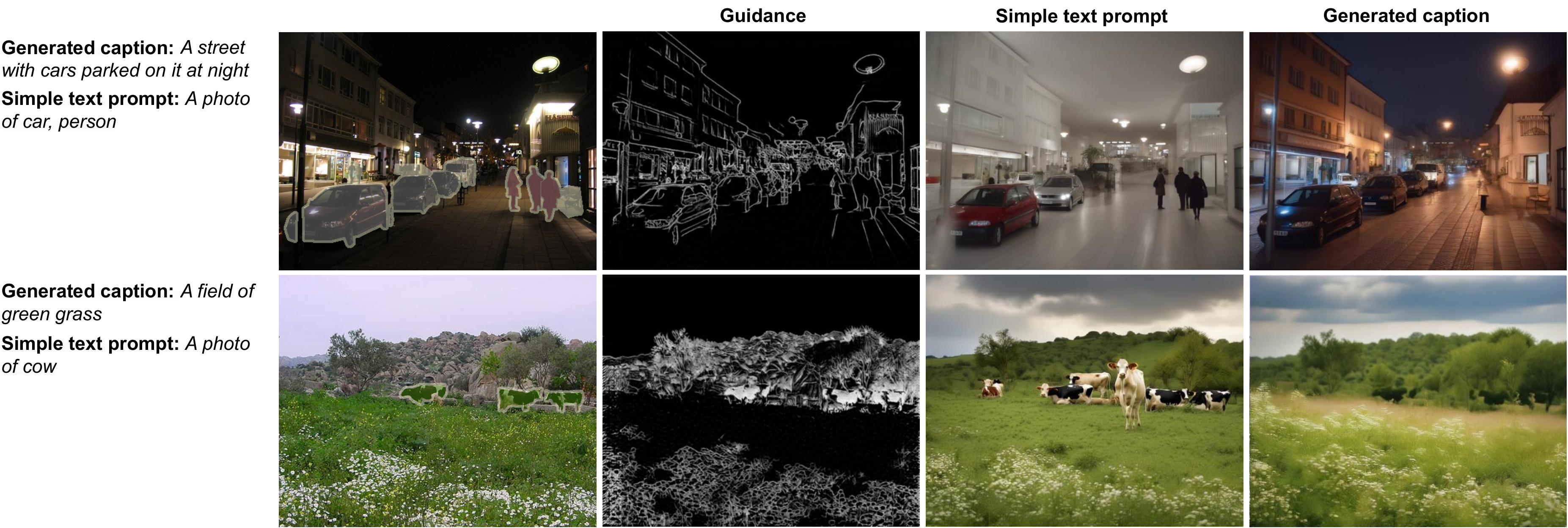}
    \caption{\textbf{Common issues of using generated captions and simple text prompts}: In addition to generated prompts and simple text prompts, the four visualizations include: original images with labeled classes, guidance images (line art), images generated by simple text prompts, and images generated by generated prompts.}
    \label{diff_prompt}
\end{figure*}

Early studies in synthetic data generation \cite{t2iadapter,controlnet,seggen1} propose using semantic segmentation maps to guide image generation and create an efficient solution for data augmentation for semantic segmentation. Labeling each pixel to show which class it belongs to helps to create accurate images with correct object locations and details. However, these methods require training on specific segmentation datasets, which limits the generation of synthetic images for classes not included in the training data. For example, \cite{t2iadapter,controlnet} propose models trained on the ADE20K dataset \cite{ade20k}, which has various images from different contexts like indoor, outdoor, industrial, and natural scenes. When generating synthetic images based on segmentation masks from the BDD100K dataset \cite{bdd}, which consists of images captured from car dashcams in self-driving scenarios, the model is unable to generate classes not present in the ADE20K dataset. These missing classes include traffic signs, traffic lights, and lane markings. Due to this limitation, we avoid using models guided by semantic segmentation maps in this study to maintain the generality of our method.

Instead of directly selecting synthetic images to train the model, some studies \cite{aug4,aug_wsss} propose using post-filtering techniques to choose the best synthetic images. However, selecting one high-quality image from many can be time-consuming, and imperfect filtering can lead to a low-quality synthetic dataset. In addition, with the ability to generate different images from the same input and change the random seeds, selecting the best ones based on multiple results does not accurately reflect the image generation ability. Therefore, in this paper, we directly use the generated images without going through any post-filtering techniques to demonstrate the effectiveness of the proposed method. In addition, in Section \ref{sec:component}, we also integrate the filter using CLIP Encoder \cite{aug4} to demonstrate the compatibility of the proposed method when applying filters.

\begin{figure*}
\centering
\begin{subfigure}{0.9\textwidth}
  \centering
  \includegraphics[width=\textwidth]{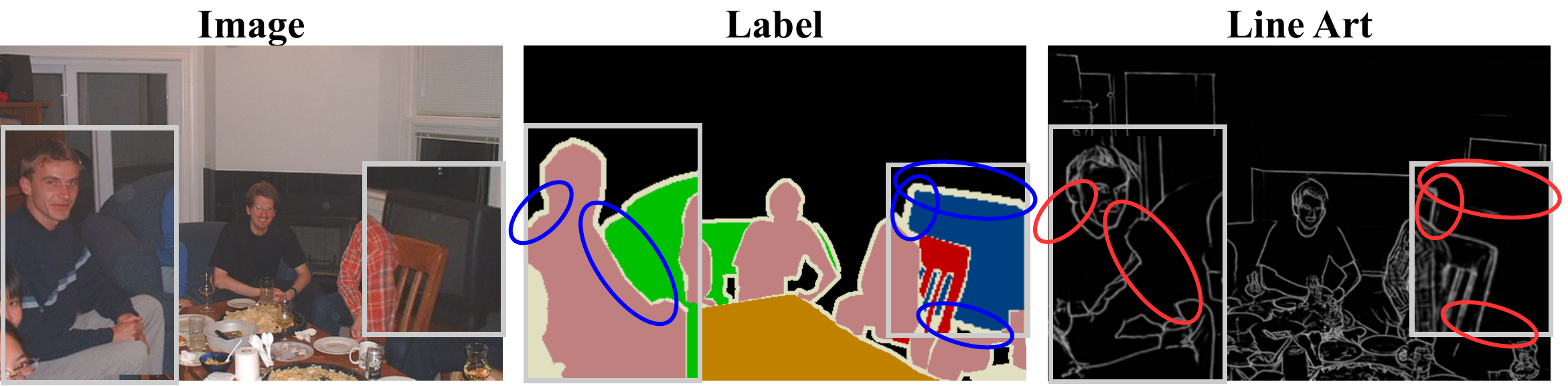}
  \caption{An example of information loss occurs when using Line Art Detection, particularly in the reddish areas, where information is lacking.}
  \label{miss}
\end{subfigure}
\vskip 0.5em
\begin{subfigure}{0.9\textwidth}
  \centering
  \includegraphics[width=\textwidth]{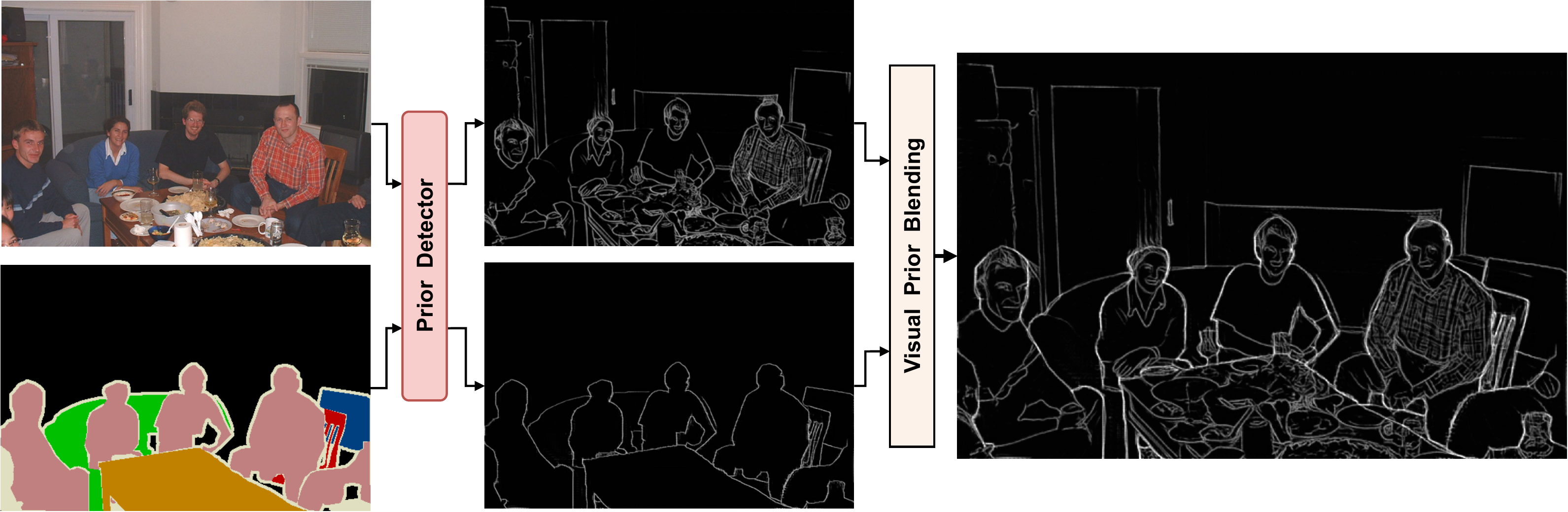}
  \caption{The results of the Visual Prior Blending method demonstrate that previously edge-deficient objects are now fully detailed, and the edge features of the labeled classes are more prominent compared to the background, thereby enhancing the generative model's attention capability.}
  \label{combine_img}
\end{subfigure}
\vskip 0.5em
\caption{The edge feature results before (a) and after (b) using the Visual Prior Blending method..}
\label{fig:fig}
\end{figure*}
\section{\uppercase{Method}}
Our proposed image data augmentation pipeline is shown in Figure \ref{main}, which consists of three main components: (1) Text prompt construction, (2) Visual Prior Blending, and (3) Controllable Diffusion Generation. \textit{Class-Prompt Appending} append “\textit{class prompt}” including the classes visible in the image with “\textit{caption}” generated from the Image Captioning model. \textit{Visual Prior Blending} method combines the visual pre-information of the real image and the segmentation map. The results of the above two methods are fed into the Controllable Stable Diffusion model to generate the synthetic image. In addition, to generate synthetic data from a given dataset, we use \textit{class balancing algorithm} to generate data with even distribution among classes. Next, we provide a detailed description of each component.

\subsection{Preliminary}

Diffusion models comprise both forward and reverse processes \cite{ddpm}. In the forward process, a Markovian chain is defined with noise added to the clean image $x_0$:

\begin{equation}
x_t = \sqrt{\bar{\alpha}_t} x_0 + \sqrt{1 - \bar{\alpha}_t} \epsilon, \quad \epsilon \sim \mathcal{N}(0, I)
\end{equation}

where $x_t$ is the noise image at time step $t$, $\epsilon$ is a noise map sampled from a Gaussian distribution, and $\bar{\alpha}_t$ denotes the corresponding noise level. The neural network $\epsilon_t$ is parameterized by $\theta$, which is optimized to predict the noise added to $\epsilon_t$ in the reverse process. A classical Diffusion model is typically optimized by:

\begin{equation}
\mathcal{L}_{DDPM} = \mathbb{E}_{x_0, t, \epsilon_t} \left\| \epsilon_t - \epsilon_{\theta}(x_t, t) \right\|_2^2
\end{equation}

% The Latent Diffusion Model (LDM) conducts the diffusion process on the latent space through a time-conditioned UNet denoted by $z_t$. The autoencoder, denoted by $\epsilon_\theta (z_t, t)$, is trained to predict the denoising variant of $z_t$. LDM has a component $y$ that controls the synthesis process as a condition (texts, semantic maps, and similar resources). Condition $y$ will be encoded before being used as input through the cross-attention layers in the intermediate layers of the UNet. This approach diversifies the input modality. The model is optimized using the following objective function:

% \begin{equation}
% \mathcal{L}_{LDM} = \mathbb{E}_{x_0, y, t, \epsilon_t} \left\| \epsilon_t - \epsilon_{\theta}(x_t, t, y) \right\|_2^2
% \end{equation}

In the context of controllable generation \cite{controlnet,t2iadapter}, when given a condition image $c_v$ and a text prompt $c_t$, the diffusion training loss function at time $t$ can be re-written as:

\begin{equation}
\mathcal{L} = \mathbb{E}_{x_0, c_v, c_t, t, \epsilon_t} \left\| \epsilon_t - \epsilon_{\theta}(x_t, t, c_v, c_t) \right\|_2^2
\end{equation}

\subsection{Generative Data Augmentation Pipeline}

\subsubsection{Text prompt construction}

To generate an image containing the labeled classes as in the original image, we need a robust prompt that describes the original image well to serve as input to the SD model. A simple way to do this is by constructing a prompt based on the labeled classes. For example, if image $\mathcal{I}_i$ contains target classes $\mathcal{C}_i$ = [$c_1$,..., $c_M$], where $M$ is the number of classes in image $\mathcal{I}_i$, we can construct a simple prompt such as: “$c_1$,..., $c_M$” or “A photo of $c_1$,..., $c_M$”. However, using a prompt that only lists target classes may not clearly describe the original image's layout. To improve this, we can use the existing or generated captions from the training images in these datasets as text prompts for SD. For example, we can use the provided captions when using the COCO dataset \cite{coco}. However, most datasets, such as PASCAL VOC \cite{voc}, BDD100K \cite{bdd}, and ADE20K \cite{ade20k}, do not have captions, while annotating captions for images also requires intensive labor. Therefore, we propose using an Image Captioning model such as BLIP-2 \cite{blip2} to generate captions for each image. However, image captions have limitations compared to simple prompts, as they often omit some of the actual classes present in the image (as shown in Figure \ref{loseclass}). This issue results in missing class names in the captions when using generated descriptions.

Illustrated in Figure \ref{diff_prompt}, some examples demonstrate using simple text prompts and generated captions to create synthetic images through the Controllable SD model. The results indicate that using simple text prompts produces images with messy layouts that do not match the original while using generated prompts leads to missing classes in the generated images due to their absence in the generated captions. However, we observe that these two methods can complement each other's weaknesses. Therefore, we propose combining generated captions with the image's class labels to address these limitations. With an image $\mathcal{I}_i$, we append the generated captions $\mathcal{P}_i^{g}$ with the class labels $\mathcal{P}_i^{c}$ to generate new text prompts $P^*_i$. This process, known as \textit{Class-Prompt Appending} \cite{datasetsd}, can be represented as: $\mathcal{P}^*_i$ = “$\mathcal{P}_i^{g}$; $\mathcal{P}_i^{c}$”. For example, in the sub-figure in the top-right corner of Figure \ref{loseclass}, the prompt generated by our proposed method would be “a pink plane on the tarmac; aeroplane, person”. Our method ensures that new text prompts include both general information and the target classes of the original image. This technique helps the synthetic image to have a clear layout similar to the original one and also addresses the problem of missing labeled classes in the synthetic image.

\newcommand\mycommfont[1]{\footnotesize\ttfamily\textcolor{blue}{#1}}
\SetCommentSty{mycommfont}
\SetKwInput{KwInput}{Input}              
\SetKwInput{KwOutput}{Output} 
\begin{algorithm}[!h]
\caption{Class balancing algorithm for dataset generation}
\label{balance_classes}
\DontPrintSemicolon
  
  \KwInput{Original dataset $\mathcal{D}_{origin}$; Target images per class $n_{balance}$}
  \KwOutput{Balanced dataset $\mathcal{D}_{final}$}
  \textbf{Stage 1: Initialization}   
  \tcc*{\fontfamily{qcr}\selectfont Storing images per class}
  $\mathcal{M} \gets \emptyset$
  
  \For{$\mathcal{I}$ in $\mathcal{D}$}
    {
    \For{$\mathcal{C}$ in $\mathcal{I}$}
        {
            $\mathcal{M}[\mathcal{C}] \gets \mathcal{M}[\mathcal{C}] \cup \{\mathcal{I}\}$
        }
    }
  \textbf{Stage 2: Sorting}   
  \tcc*{\fontfamily{qcr}\selectfont Sorting by number of classes}
    \For{$\mathcal{C}$ in $\mathcal{M}$}
    {
        Sort $\mathcal{M}[\mathcal{C}]$
    }
  \textbf{Stage 3: Balancing}   
  \tcc*{\fontfamily{qcr}\selectfont Generating additional images}
    $\mathcal{D}_{gen} \gets \emptyset$
    
    \For{$\mathcal{C}$ in $\mathcal{M}$}
    {
        \While{len($\mathcal{M}[\mathcal{C}]$) $< n_{balance}$}
       {
       	\For{$\mathcal{I}$ in $\mathcal{M}[\mathcal{C}]$}
            {
                Generate $\mathcal{I}_{gen}$ based on $\mathcal{I}$

                $\mathcal{M}[\mathcal{C}] \gets \mathcal{M}[\mathcal{C}] \cup \{\mathcal{I}_{gen}\}$
                
                $\mathcal{D}_{gen} \gets \mathcal{D}_{gen} \cup \{\mathcal{I}_{gen}\}$

                \If{len($\mathcal{M}[\mathcal{C}]$) $\geq n_{balance}$}
                {
                    \textbf{break}
                }
            }
       }
    }
$\mathcal{D}_{final} \gets \mathcal{D}_{gen} \cup \mathcal{D}_{origin}$

\Return $\mathcal{D}_{final}$
\end{algorithm}

\subsubsection{Visual Prior Blending}\label{combine_sec}

\begin{table*}[]
\caption{\textbf{Semantic Segmentation Evaluation:} Comparison in mIoU (\%) on val set between models' training on the original training set ($D_{origin})$ and the extended training set ($D_{final}$).}
\label{main_table}
% \resizebox{\textwidth}{!}{%
\centering
\begin{tabular}{lllcllcllccccc}
\toprule
\textbf{Dataset}                                                                &  &  &                                                                                 & \textbf{} &  & \textbf{VOC7}                                                                        & \textbf{} &  & \multicolumn{5}{c}{\textbf{VOC12}}                                                                                                                                                                                                                                                                                                                                                                                                                 \\ \midrule
Number images                                                                   &  &  &                                                                                 &           &  & 209                                                                                  &           &  & 92                                                                                   & 183                                                                                  & 366                                                                                  & 732                                                                                  & 1464                                                                                   \\ \midrule
\multirow{2}{*}{\begin{tabular}[c]{@{}l@{}}DeepLabV3+\\ Resnet50\end{tabular}}  &  &  & $D_{origin}$                                                                    &           &  & 46.54                                                                                &           &  & 29.91                                                                                & 38.21                                                                                & 49.40                                                                                & 58.20                                                                                & 61.84                                                                                  \\ \cmidrule{2-14} 
                                                                                &  &  & \begin{tabular}[c]{@{}c@{}}$D_{gen} \cup D_{origin}$\\ $\triangle$\end{tabular} &           &  & \begin{tabular}[c]{@{}c@{}}50.27\\ \color[HTML]{3531FF} $\uparrow$ 3.73\end{tabular} &           &  & \begin{tabular}[c]{@{}c@{}}33.87\\ \color[HTML]{3531FF} $\uparrow$ 3.96\end{tabular} & \begin{tabular}[c]{@{}c@{}}41.45\\ \color[HTML]{3531FF} $\uparrow$ 3.24\end{tabular} & \begin{tabular}[c]{@{}c@{}}52.22\\ \color[HTML]{3531FF} $\uparrow$ 2.78\end{tabular} & \begin{tabular}[c]{@{}c@{}}60.11\\ \color[HTML]{3531FF} $\uparrow$ 1.91\end{tabular} & \begin{tabular}[c]{@{}c@{}}63.06\\ \color[HTML]{3531FF} $\uparrow$ 1.22\end{tabular}   \\ \midrule
\multirow{2}{*}{\begin{tabular}[c]{@{}l@{}}PSPNet\\ Resnet50\end{tabular}}      &  &  & $D_{origin}$                                                                    &           &  & 47.04                                                                                &           &  & 31.87                                                                                & 38.96                                                                                & 46.62                                                                                & 57.48                                                                                & 62.39                                                                                  \\ \cmidrule{2-14} 
                                                                                &  &  & \begin{tabular}[c]{@{}c@{}}$D_{gen} \cup D_{origin}$\\ $\triangle$\end{tabular} &           &  & \begin{tabular}[c]{@{}c@{}}50.01\\ \color[HTML]{3531FF} $\uparrow$ 2.97\end{tabular} &           &  & \begin{tabular}[c]{@{}c@{}}34.67\\ \color[HTML]{3531FF} $\uparrow$ 2.80\end{tabular} & \begin{tabular}[c]{@{}c@{}}41.46\\ \color[HTML]{3531FF} $\uparrow$ 2.50\end{tabular} & \begin{tabular}[c]{@{}c@{}}49.34\\ \color[HTML]{3531FF} $\uparrow$ 2.72\end{tabular} & \begin{tabular}[c]{@{}c@{}}61.09\\ \color[HTML]{3531FF} $\uparrow$ 3.61\end{tabular} & \begin{tabular}[c]{@{}c@{}}63.78\\ \color[HTML]{3531FF} $\uparrow$ 1.39\end{tabular}   \\ \midrule
\multirow{2}{*}{\begin{tabular}[c]{@{}l@{}}Mask2Former\\ Resnet50\end{tabular}} &  &  & $D_{origin}$                                                                    &           &  & 48.28                                                                                &           &  & 34.85                                                                                & 39.63                                                                                & 51.37                                                                                & 59.94                                                                                & 63.65                                                                                  \\ \cmidrule{4-14} 
                                                                                &  &  & \begin{tabular}[c]{@{}c@{}}$D_{gen} \cup D_{origin}$\\ $\triangle$\end{tabular} &           &  & \begin{tabular}[c]{@{}c@{}}49.69\\ \color[HTML]{3531FF} $\uparrow$ 1.41\end{tabular} &           &  & \begin{tabular}[c]{@{}c@{}}35.53\\ \color[HTML]{3531FF} $\uparrow$ 0.68\end{tabular} & \begin{tabular}[c]{@{}c@{}}40.29\\ \color[HTML]{3531FF} $\uparrow$ 0.66\end{tabular} & \begin{tabular}[c]{@{}c@{}}51.77\\ \color[HTML]{3531FF} $\uparrow$ 0.40\end{tabular} & \begin{tabular}[c]{@{}c@{}}60.02\\ \color[HTML]{3531FF} $\uparrow$ 0.08\end{tabular} & \begin{tabular}[c]{@{}c@{}}62.56\\ \color[HTML]{FE0000} $\downarrow$ 1.09\end{tabular} \\ \midrule
\end{tabular}%
\end{table*}

Unlike Stable Diffusion models that typically only use text prompts to generate images, Controllable Generation models require additional input guidance (Canny Edge \cite{canny}, Sketch-Guided \cite{sketch}, Line-Art Edge \cite{lineart}, Depth Map \cite{depth}, HED soft edge \cite{hed}) generated from the visual prior detector to determine the image layout. Canny, Line-Art Edge, and Sketch are the visual priors we propose using to balance the diversity of the image and the precise structure of the generated classes in the image. Our default visual prior is Line-Art Edge to generate visual guidance. In Section \ref{priors_sec}, we also discuss other types of visual priors to see how effectively they augment the data. Using the T2I-Adapter \cite{t2iadapter} model, the Line Art Detector converts the image into visual prior in-line drawings for each input image. Then, the Adapter generates different resolution features, performing conditional operations at each time step with the UNet denoiser's features.

In general, methods such as Line-Art Edge, Canny Edge, or HED soft edge all suffer from the limitation that the labeled classes in the image may be blurred or small in size, leading to inaccuracies in describing the structure of the labeled classes within the conditional image. Figure \ref{miss} shows an image produced using Line Art Detection. However, the edge results in this case are missing some details of the person, and the TV monitor is almost absent. The red shapes indicate the missing details in the image. This loss of information also occurs when using HED or Canny Edge. These weaknesses result in mislabeling in the synthetic image compared to the original image. We observed that although the segmentation labels of real images cannot fully describe an image's content, they provide accurate information about the labeled classes. Based on this observation, we propose combining the real image's prior visualization with the labels before feeding them into the controlled image generation model. The blending of the prior visual image $\mathcal{I}_i$ ($\mathcal{V}^\mathcal{I}_i$) and prior visual segmentation label $\mathcal{S}_i$ ($\mathcal{V}^\mathcal{S}_i$) ensures that the generated image has a clear layout and well preserves structures the class labeled information. Our proposed blends $\mathcal{V}^\mathcal{I}_i$ and $\mathcal{V}^\mathcal{S}_i$ by a weighted sum:

\begin{equation}
    \label{combine}
    \mathcal{V}^*_i = \omega_1 \mathcal{V}^\mathcal{I}_i + \omega_2 \mathcal{V}^\mathcal{S}_i
\end{equation}

With $\omega_1$, $\omega_2$ being the trade-off scales when combining $\mathcal{V}^\mathcal{I}_i$ and $\mathcal{V}^\mathcal{S}_i$. This blending results in a prior visual that is clear in content and complete information about labeled classes. Figure \ref{combine_img} shows how the Visual Prior Blending method can preserve the structure labeled classes in an image.

\subsection{Create class-balancing dataset}

To address the issue of class imbalance during model training, we aim for the final dataset $D_{final}$, which merges the original dataset $D_{origin}$ and the synthetic dataset $D_{gen}$, to have a balanced distribution among classes. To create a balanced dataset from the original dataset $D_{origin}$, we use the class balancing algorithm to generate the dataset $D_{gen}$ based on the balancing factor $n_{balance}$, as presented in Algorithm \ref{balance_classes}. The algorithm consists of three main stages: Initialization, Sorting, and Balancing. In Stage 1, a dictionary $\mathcal{M}$ is initialized to map each class to its associated images. Each image $\mathcal{I}$ is linked to a list of classes it contains. In the next stage, images are arranged in ascending order based on the number of classes they represent, prioritizing those with fewer classes to be generated to maintain the balance. In the final stage, additional images are generated for each class until they reach $n_{balance}$, ensuring an even distribution among classes. This process ensures the dataset is balanced, preventing the overrepresentation of any class and promoting more robust model training.

After generating high-quality training samples, the synthetic dataset $D_{gen}$ and the original dataset $D_{origin}$ are merged into an extended dataset $D_{final}$ for training:

\begin{equation}
    D_{final} = D_{gen}  \cup D_{origin}
\end{equation}

In the default setting, we choose an appropriate $n_{balance}$ such that $|D_{gen}| \approx |D_{origin}|$ where $|$  $.$  $|$ is the number of images in the dataset. In Section \ref{effect_number}, we further discuss the impact of varying the number of generated synthetic images.

\section{\uppercase{Results}}

\subsection{Experiment Details}
\subsubsection{Dataset}
In this section, we evaluate our method on the segmentation datasets VOC7 and VOC12 \cite{voc}. PASCAL VOC 2007 has 422 images annotated for semantic segmentation, split into 209 training and 213 validation images. Meanwhile, VOC12 has training and validation sets, including 1.464 and 1.449 images. In addition to training on the entire VOC12 dataset (1.464 images), we also train our model using 1/2 (732 images), 1/4 (366 images), 1/8 (183 images), and 1/16 (92 images) partition protocols \cite{u2pl}. These evaluations on smaller subsets demonstrate the effectiveness of our method in real-world, limited-data scenarios.
 
\subsubsection{Implementation details} \label{implement}
We construct our framework on the deep learning PyTorch framework \cite{pytorch} and T2I-Adapter \cite{t2iadapter} using Stable Diffusion XL 1.0 \cite{SDXL} with 30 time steps. We generate data using values for $\omega_1$ and $\omega_2$, as defined in Section \ref{sec:scales}. For semantic segmentation, we employ the DeepLabV3+ \cite{deeplabv3plus}, PSPNet \cite{pspnet}, and Mask2Former \cite{mask2former} with segmenters implemented in the MMSegmentation framework \cite{mmseg}. We utilize the SGD optimizer with standard settings in MMSegmentation. We train our models with an input image size of 512$\times$512 with 30k steps on the Pascal VOC datasets, including VOC7 and VOC12. During training, we only apply simple transformation methods to augment data, such as: \textit{RandomResize, RandomCrop, RandomFlip}. The performance of the trained model on the applied data is assessed using the Mean Intersection over Union (mIoU) metric.

\subsection{Main Results}

\subsubsection{Quantitative results:}

\begin{table*}[]
\centering
\caption{Evaluation results of the DeepLabV3+ model on the PASCAL VOC7 dataset trained on $D_{origin}$ and $D_{final}$.}
\label{each_class}
\resizebox{1.8\columnwidth}{!}{%
\begin{tabular}{llcccccccccc}
\toprule
            & \includegraphics[width=20px]{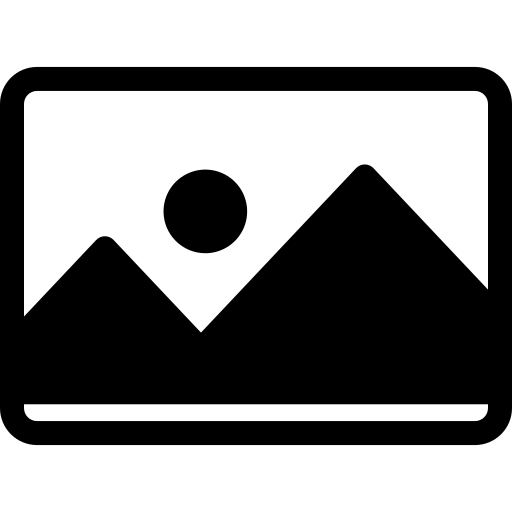}                       & \includegraphics[width=16px]{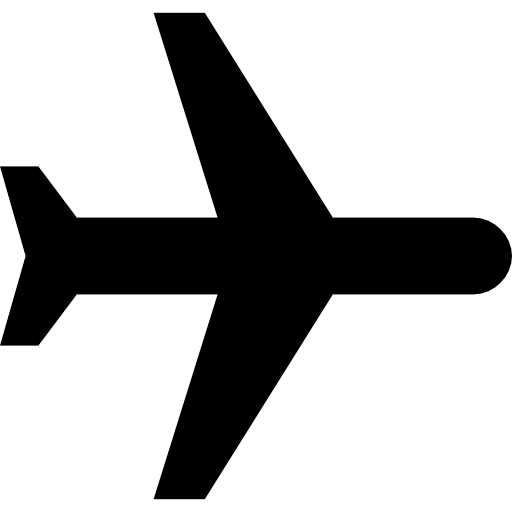}            & \includegraphics[width=18px]{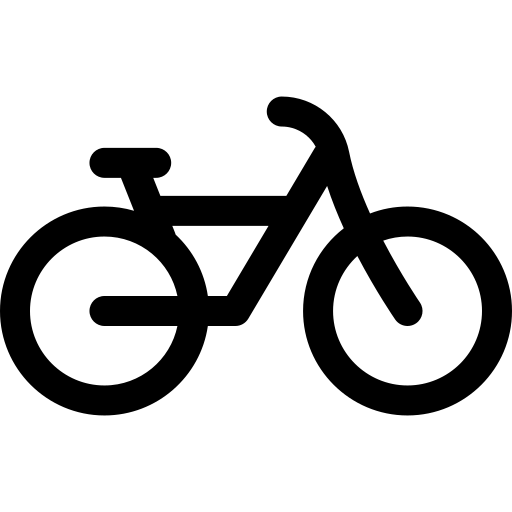}              & \includegraphics[width=18px]{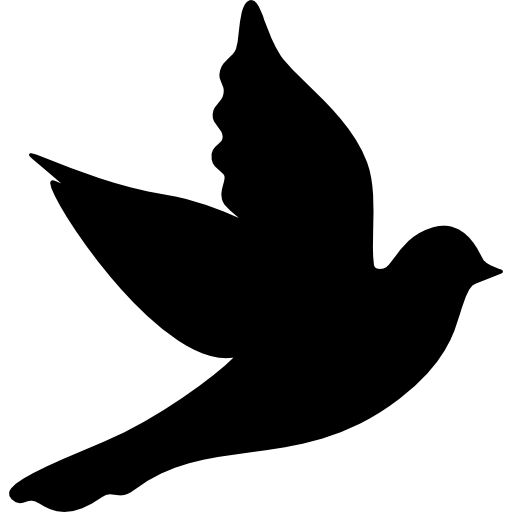}                 & \includegraphics[width=18px]{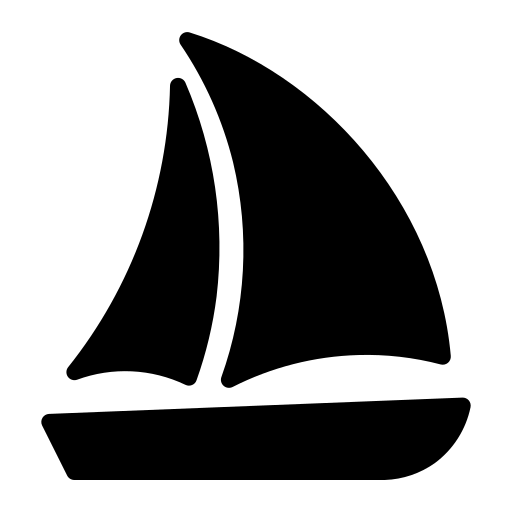}                 & \includegraphics[width=18px]{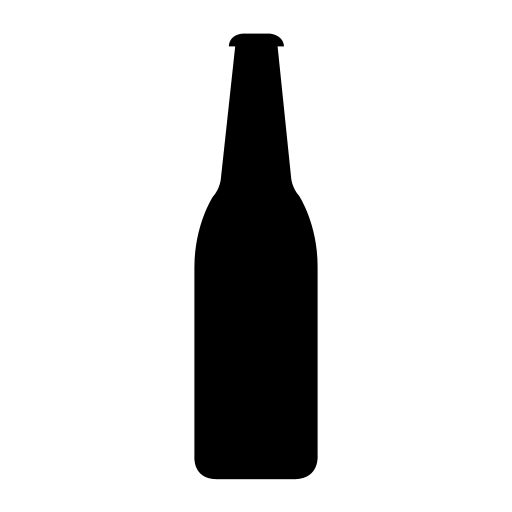}               & \includegraphics[width=18px]{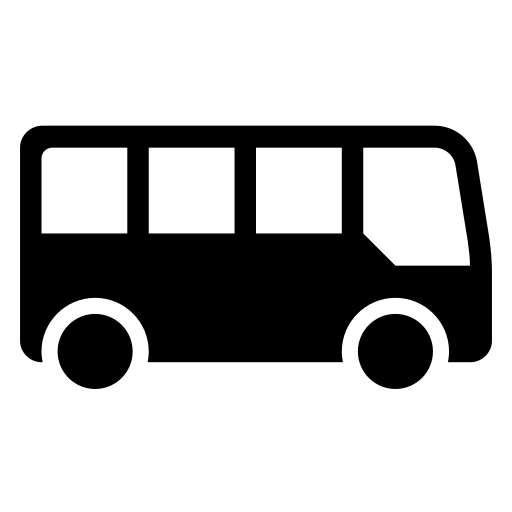}                  & \includegraphics[width=18px]{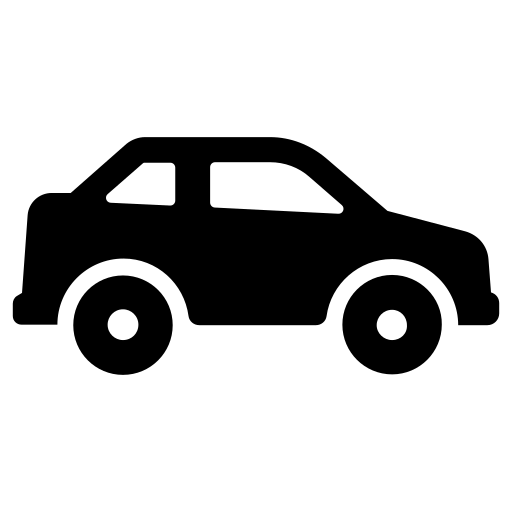}                  & \includegraphics[width=18px]{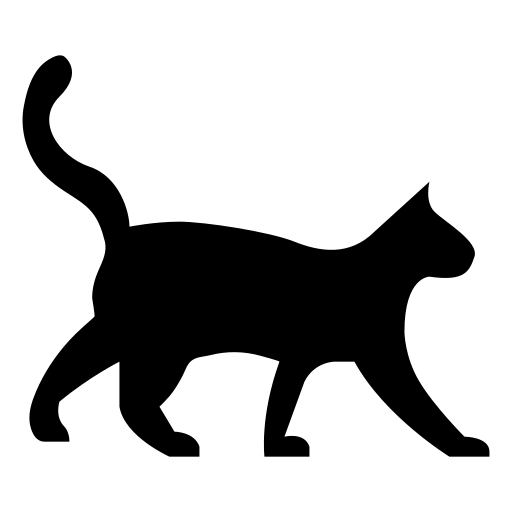}                  & \includegraphics[width=18px]{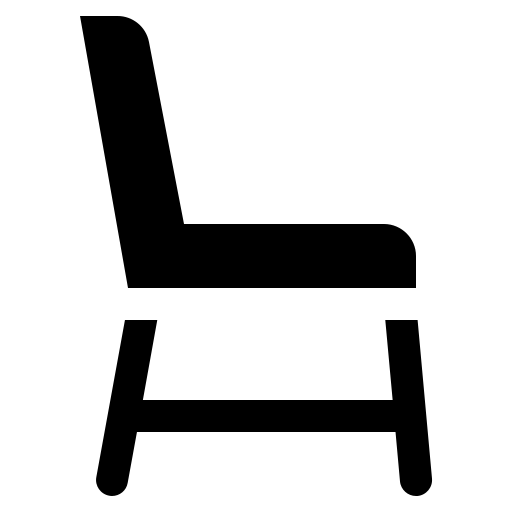}                & \includegraphics[width=18px]{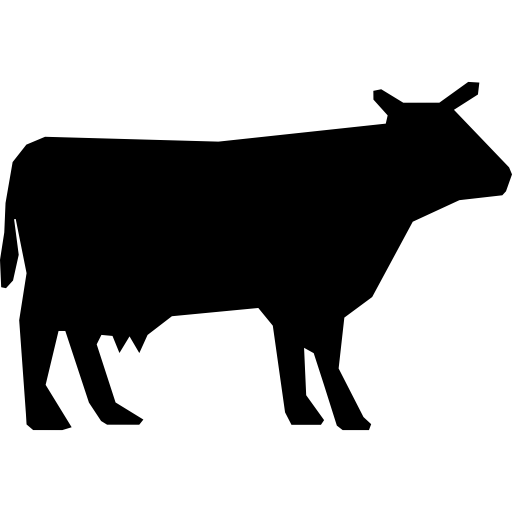}                  \\ \midrule
$D_{origin}$   & \textbf{89.32}                   & \textbf{75.35}       & 43.90                & 53.95                & 38.03                & 51.72                & 50.81                & 67.41                & 72.0                 & 7.77                 & 24.61                \\ \midrule
$D_{final}$ & 88.69                            & 67.67                & \textbf{45.35}       & \textbf{57.00}       & \textbf{38.07}       & \textbf{56.08}       & \textbf{70.14}       & \textbf{70.67}       & \textbf{73.46}       & \textbf{28.81}       & \textbf{45.45}       \\ \bottomrule
            &                                  & \multicolumn{1}{l}{} & \multicolumn{1}{l}{} & \multicolumn{1}{l}{} & \multicolumn{1}{l}{} & \multicolumn{1}{l}{} & \multicolumn{1}{l}{} & \multicolumn{1}{l}{} & \multicolumn{1}{l}{} & \multicolumn{1}{l}{} & \multicolumn{1}{l}{} \\ \toprule
            & \multicolumn{1}{c}{\includegraphics[width=16px]{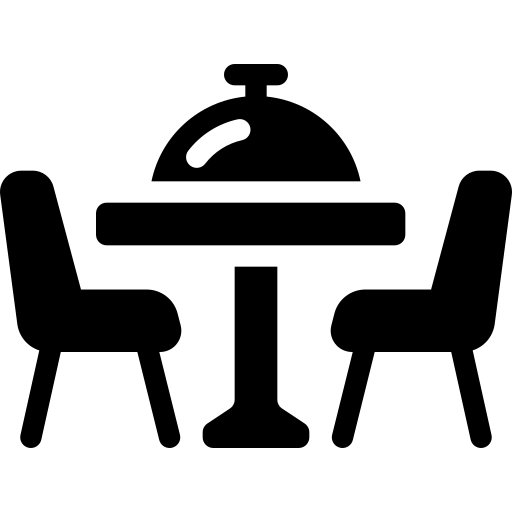}} & \includegraphics[width=18px]{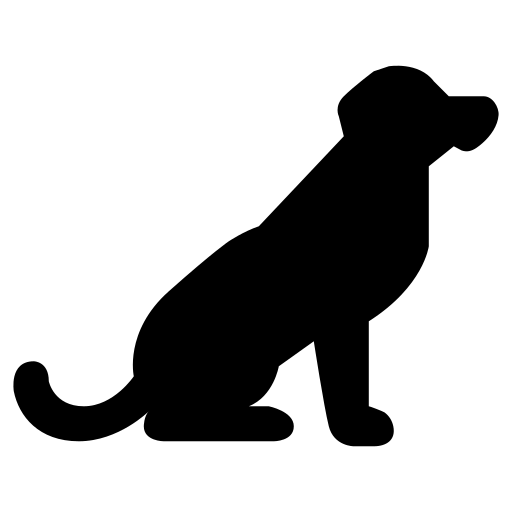}                  & \includegraphics[width=20px]{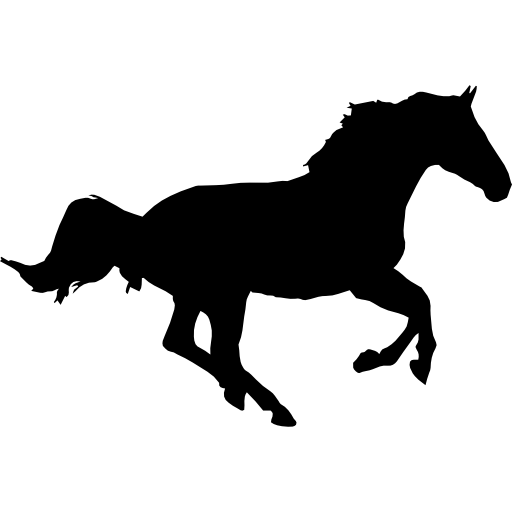}                & \includegraphics[width=18px]{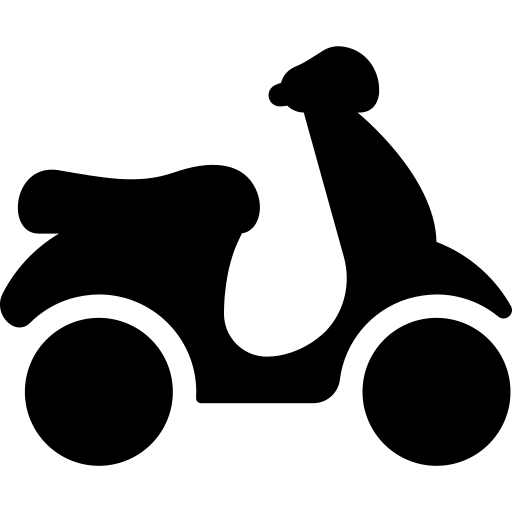}            & \includegraphics[width=18px]{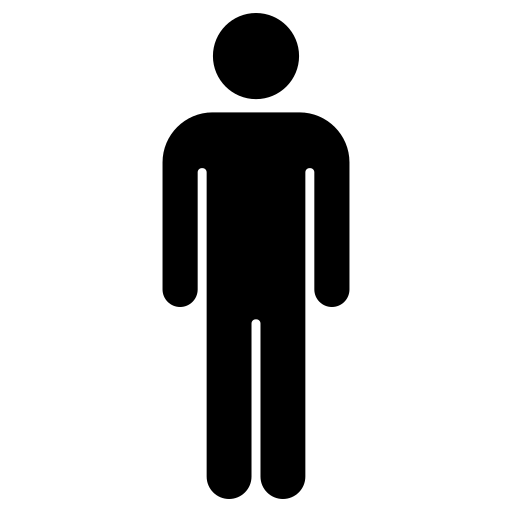}               & \includegraphics[width=18px]{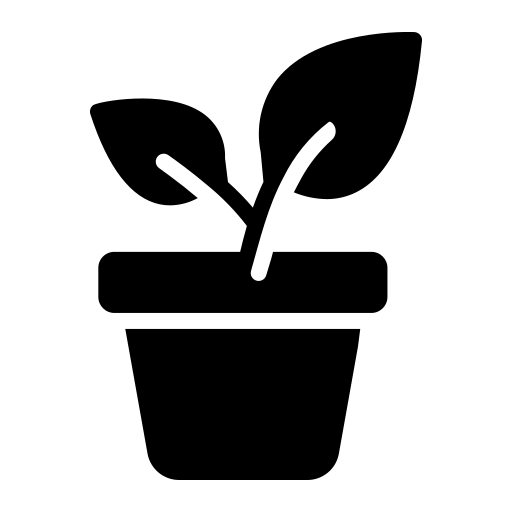}         & \includegraphics[width=18px]{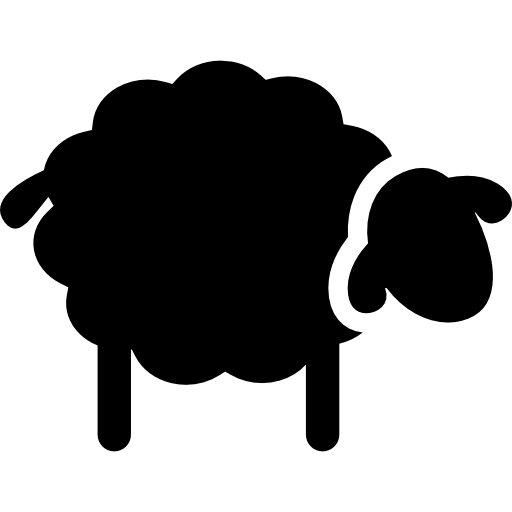}                & \includegraphics[width=17px]{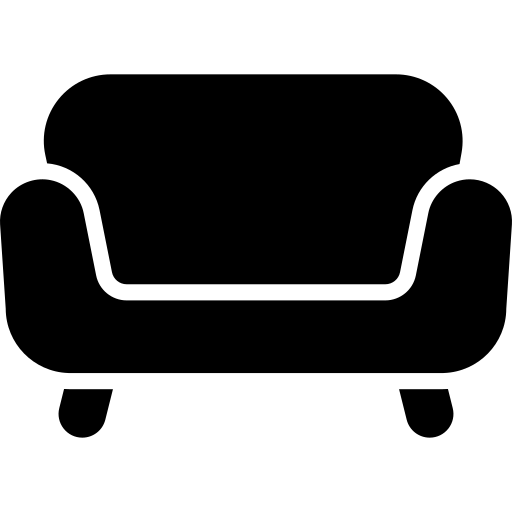}                 & \includegraphics[width=18px]{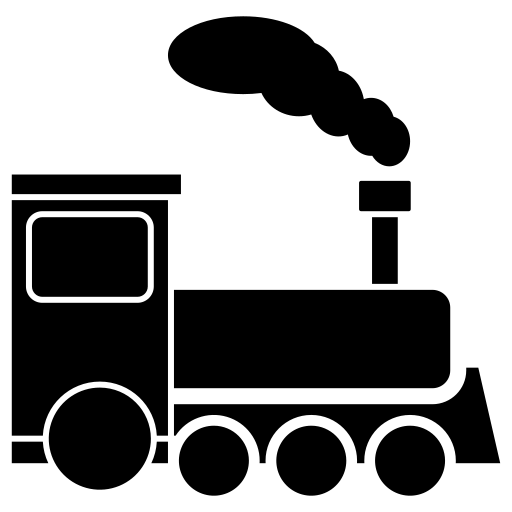}                & \includegraphics[width=18px]{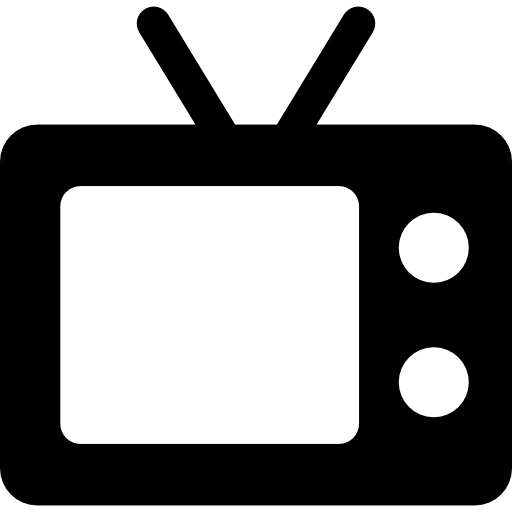}           & \textbf{mIoU (\%)}            \\ \midrule
$D_{origin}$   & \multicolumn{1}{c}{\textbf{41.67}}        & \textbf{45.16}                & 42.29                & \textbf{66.08}                & \textbf{72.71}                & \textbf{42.00}                & 14.07                & \textbf{24.68}                & 40.74                & 12.96                & 46.54                \\ \midrule
$D_{final}$ & \multicolumn{1}{c}{39.48}        & 36.27                & \textbf{50.10}       & 55.80                & 71.49                & 40.29                & \textbf{21.91}       & 18.10       & \textbf{48.13}       & \textbf{32.72}       & \textbf{50.27}       \\ \bottomrule
\end{tabular}%
}
\end{table*}

\begin{figure}[!t]
  \centering
    \includegraphics[width=0.45\textwidth]{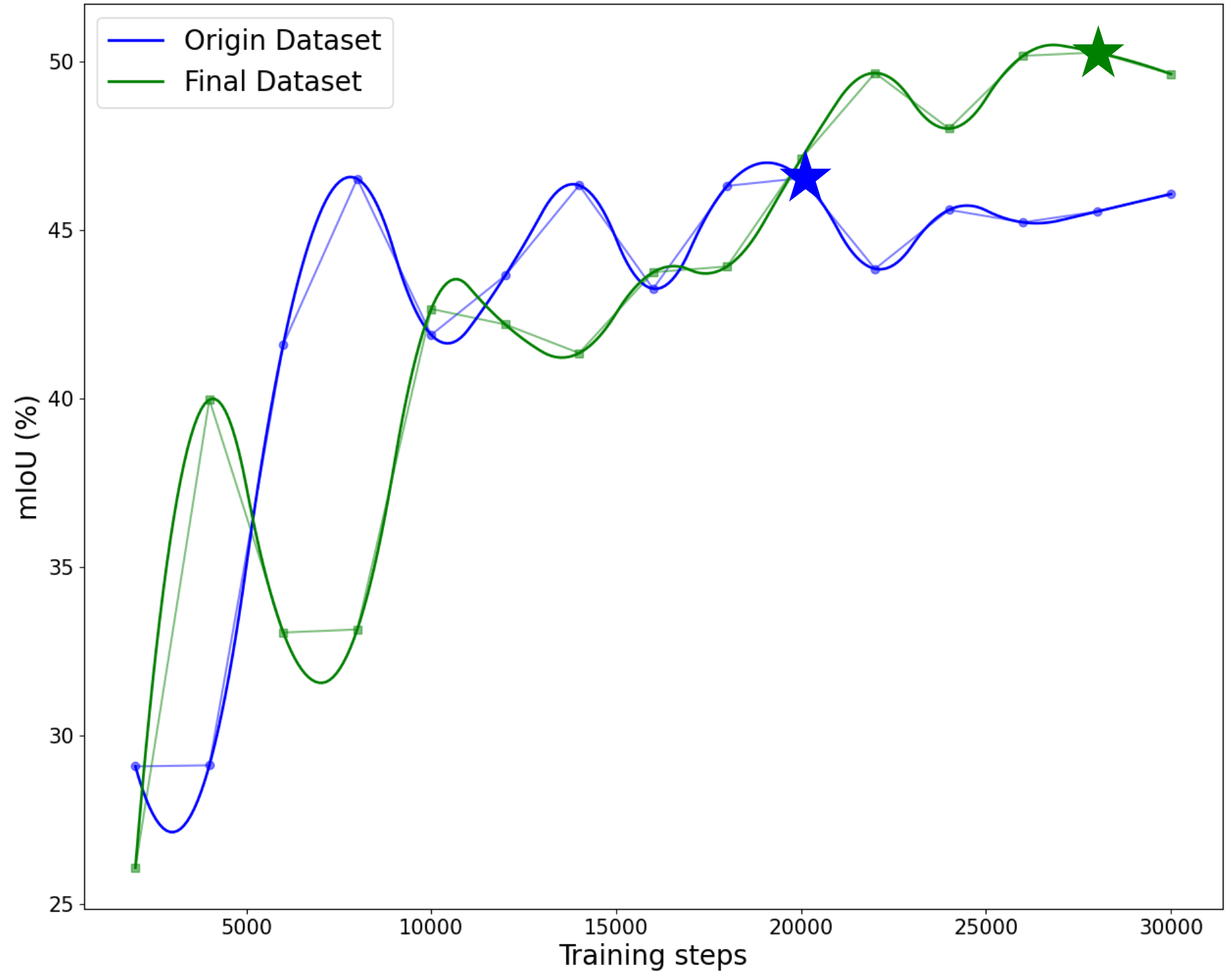}
    \caption{The mIoU (\%) of the DeepLabV3+ model over 30k training steps. The star symbol indicates the point of convergence, where the model achieves its highest performance on the validation set.}
    \label{train_steps}
\end{figure}

The results presented in Table \ref{main_table} demonstrate that combining augmented data ($D_{gen}$) with the original dataset ($D_{origin}$) improves the performance of various segmentation models on the VOC7 and VOC12 datasets. All models, including DeepLabV3+, PSPNet, and Mask2Former, significantly improve when augmented data. DeepLabV3+ consistently performs the best across different dataset sizes. We note that as the amount of real-world data increases, the accuracy of the generated images and labels becomes crucial. Mismatched generated images in the synthetic data can lead to performance degradation; this is observed in Mask2Former when trained on the VOC12 dataset with 1464 images.
\begin{figure*}
    \centering
    \includegraphics[width=0.91\textwidth]{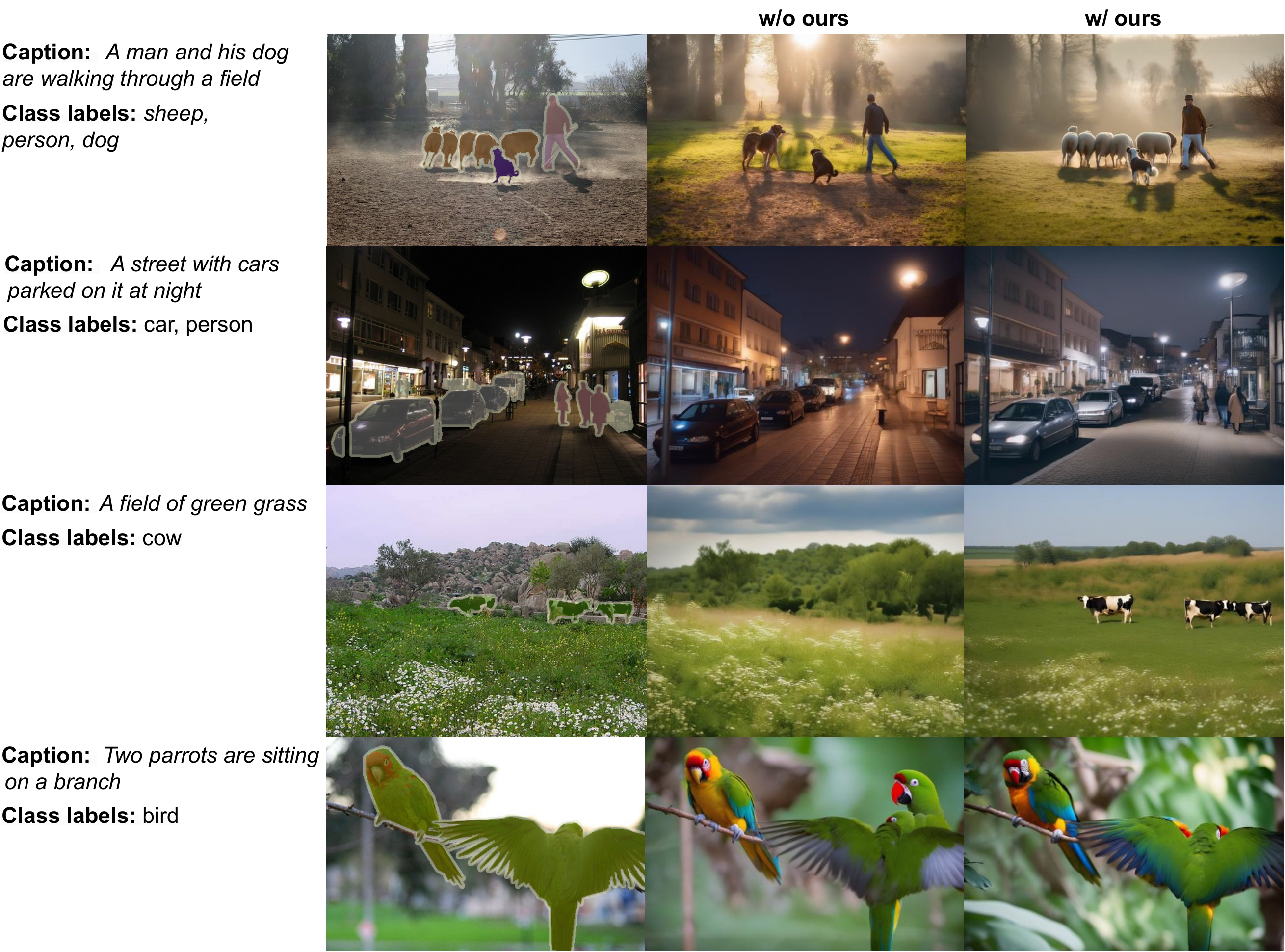}
    \caption{Some results show the limitations of direct generation and the effectiveness of our method to overcome them.}
    \label{example}
\end{figure*}

\begin{table}[]
\centering

\caption{Impact of Class-Prompt Appending (1), Visual Prior Blending (2), Class balancing algorithm (3), and Post Filter \cite{aug4} (4).}
\label{component}
% \resizebox{\textwidth}{!}{%
\begin{tabular}{clclclclc}
\hline
\textbf{(1)} & \textbf{} & \textbf{(2)} & \textbf{} & \textbf{(3)}         & \textbf{} & \textbf{(4)}         & \textbf{} & \textbf{mIoU (\%)} \\ \hline
             &           &              &           &                      &           &                      &           & 42.15              \\
$\checkmark$ &           &              &           &                      &           &                      &           & 47.60              \\
             &           & $\checkmark$ &           &                      &           &                      &           & 47.25              \\
$\checkmark$ &           & $\checkmark$ &           & \multicolumn{1}{l}{} &           & \multicolumn{1}{l}{} &           & 48.98              \\
$\checkmark$ &           & $\checkmark$ &           & $\checkmark$         &           & \multicolumn{1}{l}{} &           & 50.27              \\
$\checkmark$ &           & $\checkmark$ &           & $\checkmark$         &           & $\checkmark$         &           & 52.23              \\ \hline
\end{tabular}%
% }
\end{table}

\begin{table}[]
\centering
\caption{Effect of the number of synthetic data generated on data balance and performance.}
\label{number_impact}
\begin{tabular}{clccc}
\hline
\multicolumn{1}{l}{\textbf{n$_{gen}$}} & \textbf{R/S}                & \textbf{Entropy $\uparrow$} & \textbf{ClR $\downarrow$} & \textbf{mIoU (\%)} \\ \hline
-                                      & 209/0                       & 3.944                       & 0.253                     & 46.54                         \\
27                                     & 209/216                     & 4.044                       & 0.231                     & 50.27                         \\
41                                     & 209/425                     & 4.042                       & 0.231                     & \textbf{50.37}                \\
55                                     & \multicolumn{1}{c}{209/634} & \textbf{4.059}              & \textbf{0.224}            & 49.25                         \\ 
69                                     & \multicolumn{1}{c}{209/845} & 4.057              & 0.225            & 47.32                         \\ \hline
\end{tabular}%
\end{table}

Figure \ref{train_steps} visualizes the validation set accuracy on the PASCAL VOC7 dataset over 30,000 steps using the DeepLabV3+ model. The results show that training the model on the $D_{final}$ achieves 50.27\% mIoU, 3.73\% higher than training solely on the $D_{origin}$. The visualization also indicates that the model converges earlier when trained on the $D_{origin}$ compared to the $D_{final}$. Additionally, the detailed performance for each class in the PASCAL VOC dataset provided in Table \ref{each_class} demonstrates that most classes show improved accuracy when trained on the $D_{final}$ dataset. Notably, some classes, such as \textsf{``bus"}, \textsf{``cow"}, \textsf{``chair"}, and \textsf{``TV monitor"}, exhibit significant improvements.

\subsubsection{Qualitative results}

In Figure \ref{example}, each row presents three images: the original image, the image generated by the Controllable Generation model (using prompts generated by BLIP-2), and the image produced by the generation model with our proposed. For the images generated by the SD model without our method, the first two rows illustrate cases where labeled classes in the image are missing in the generated description, causing the model to fail in accurately generating all those classes. In the third row, the image contains objects belonging to the \textsf{``cow"} class with smaller sizes, which prevents the Image Captioning Model from including \textsf{``cow"} in the description, resulting in an inaccurate generated image without cows. In the last row, although the case is relatively simple and the prompt includes all the labeled classes, the generated image still fails to depict the structure of both \textsf{``bird"} objects accurately.
All four images are significantly improved with our proposed pipeline, with the labeled objects entirely generated and their structure well preserved. This demonstrates that our method effectively addresses the limitations of directly using Controllable Image Generation models.

\subsection{Ablation Study}\label{ablation}

We conduct all ablation study experiments using the DeepLabV3+ model, backbone Resnet50 with training and evaluation details described in Section \ref{implement}. The model is trained on the PASCAL VOC7 dataset, and the results are evaluated on the val set.

\subsubsection{Effects of different methods:} \label{sec:component}

The performance of the proposed methods is summarized in Table \ref{component}. When using the Baseline with generated prompts, the performance is only as low as 42.15\% mIoU, which is lower than when training on the original data. This demonstrates the essential nature of the proposed methods when generating synthetic data. Our method yields a 50.27\% mIoU, showing its effectiveness. Additionally, we experimented with combining the Post Filter with the Category-Calibrated CLIP Rank \cite{aug4} for generating synthetic data. The result of this blending is higher, which shows that our method can combine filters from previous studies \cite{aug4,aug_wsss} to improve the performance.

\subsubsection{Effect of number of synthetic data:} \label{effect_number}

\begin{table*}[]
\centering
\caption{Performance of different text prompt selections}
\label{prompt_selection}
\begin{tabular}{lllllllc}
\toprule
\textbf{Method}        &  &  &  & \textbf{Example}                                                                                            &  &  & \textbf{mIoU (\%)} \\ \midrule
Generated caption      &  &  &  & A room with a table and a laptop on it                                                                      &  &  & 47.24              \\
                       &  &  &  &                                                                                                             &  &  &                    \\
Simple text prompt     &  &  &  & A photo of sofa, chair, dining table                                                                        &  &  & 48.21              \\
                       &  &  &  &                                                                                                             &  &  &                    \\
Class-Prompt Appending &  &  &  & \begin{tabular}[c]{@{}l@{}}A room with a table and a laptop on it;\\ sofa, chair, dining table\end{tabular} &  &  & \textbf{50.27}     \\ \bottomrule
\end{tabular}%
% }
\end{table*}

\begin{table}[]
% \resizebox{\textwidth}{!}{%
\centering
\caption{Study on different trade-off scales}
\label{scales}
\begin{tabular}{cccccl}
\toprule
\multirow{2}{*}{$\boldsymbol{\omega_2}$} & \multicolumn{5}{c}{$\boldsymbol{\omega_1}$}                                                    \\ \cmidrule{2-6} 
                                         & \textbf{0.6} & \textbf{0.7}   & \textbf{0.8} & \textbf{0.9} & \multicolumn{1}{c}{\textbf{1.0}} \\ \midrule
\textbf{0.7}                             & 47.12        & 47.32          & 46.42        & 45.36        & 45.38                            \\
\textbf{0.8}                             & 48.51        & 48.93          & 48.32        & 47.09        & 46.79                            \\
\textbf{0.9}                             & 49.13        & \textbf{50.27} & 49.11        & 48.79        & 48.36                            \\
\textbf{1.0}                             & 48.13        & 49.32          & 48.91        & 48.33        & 47.03                            \\ \bottomrule
\end{tabular}%
% }
\end{table}

The results of the proposed method's experiments with different amounts of generated synthetic data are summarized in Table \ref{number_impact}, \textbf{R/S} refers to the number of real/synthetic images. The metrics we use to evaluate data imbalance include Entropy, and Class Imbalance Ratio (CIR). Initially, without using synthetic data, the model achieves an mIoU of 46.54. When training with synthetic data in quantities of approximately $|D_{origin}|$, $2 \times |D_{origin}|$, and $3 \times |D_{origin}|$, the data balance metrics stabilize at a better level, and model performance improves. However, we observed that using synthetic data around $3 \times |D_{origin}|$ negatively impacts performance, resulting in a decrease compared to training with $1 \times |D_{origin}|$ and $2 \times |D_{origin}|$ of synthetic data.

\subsubsection{Text prompt selection:} \label{promts}

Table \ref{prompt_selection} compares different text prompt selection methods for generative modeling. We compare the performance of three prompt types: \textit{Generated caption generated} from the Image Captioning model, \textit{Simple text prompt} listing the classes in the image, and \textit{Class-Prompt Appending}, a combination of the two prompt types. \textit{Class-Prompt Appending} outperforms the other two methods by 50.27 mIoU (\%), precisely 3.03 and 2.06 better than \textit{generated caption} and \textit{simple text prompt}, respectively, in mIoU. These results show that the \textit{Class-Prompt Appending} text prompt selection method can support SD in generating diverse datasets and ensuring accurate attention.

\begin{figure*}
    \includegraphics[width=\textwidth]{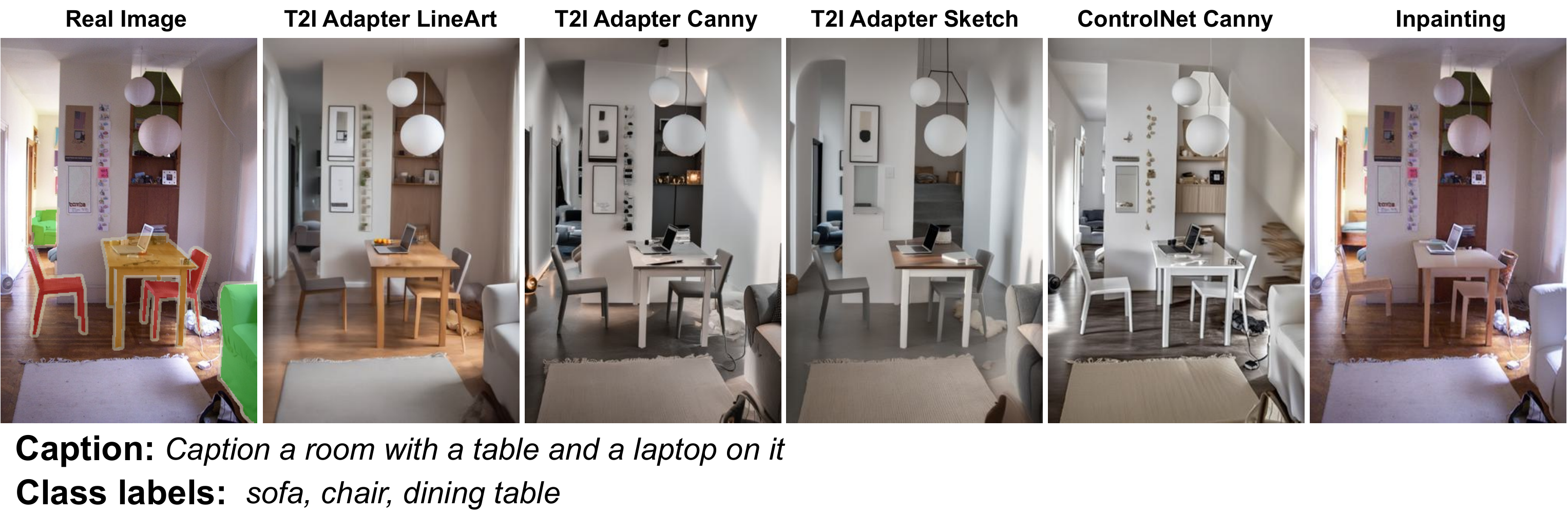}
    \caption{Some image generation results from various Controllable models when combined with our proposed approach.}
    \label{example_prior}
\end{figure*}

\begin{table*}[]
% \resizebox{\textwidth}{!}{%
\centering
\caption{Different visual priors controlled the enhancement results. All used the Stable Diffusion XL version.}
\label{priors}
\begin{tabular}{lcclclclc}
\toprule
\multirow{2}{*}{}  & \multicolumn{4}{c}{\textbf{T2I-Adapter}}               & \textbf{} & \textbf{ControlNet} &  & \multirow{2}{*}{\textbf{Inpainting}} \\ \cmidrule{2-5} \cmidrule{7-7}
                   & \textbf{LineArt} & \textbf{Canny} &  & \textbf{Sketch} & \textbf{} & \textbf{Canny}      &  &                                      \\ \midrule
\textbf{mIoU (\%)} & \textbf{50.27}            & 48.95          &  & 47.52           &           & 48.95               &  & 47.56                                \\ \bottomrule
\end{tabular}%
% }
\end{table*}

\subsubsection{Effect of different trade-off scales:} \label{sec:scales}
Trade-off scales are utilized to blend the visual prior of the image with the semantic segmentation map presented in Section \ref{combine_sec}. We tested various scales and documented the results in Table \ref{scales}. The outcomes indicate that the scale $\omega_1$=0.7 and $\omega_2$=0.9 yields the best results, with $\omega_2$ enabling proper localization of the labeled classes. On the other hand, the scale $\omega_1$=0.7 retains the general content of the image without needing to be as detailed as the original image.

\subsubsection{Other Visual Priors}\label{priors_sec}

We compare using different visual priors for both T2I-Adapter \cite{t2iadapter} and ControlNet \cite{controlnet}: Line Art \cite{lineart}, Canny \cite{canny}, Sketch \cite{sketch}. We also combine Inpainting \cite{SD} with our methods (excluding Visual Prior Blending). Although the T2I-Adapter combined with Line Art gives the best result at 50.27\% mIoU, other visual priors also show competitive performance, especially Canny on both T2I-Adapter and ControlNet. Figure \ref{example_prior} shows some image results with Controllable Diffusion model.

\section{\uppercase{Discussion and Conclusion}}
\subsection{Limitations}
While our approach effectively generates synthetic images to augment data for semantic segmentation, there are certain limitations to consider. First, the results in Tables \ref{main_table} and \ref{number_impact} show that the model's performance may decrease when the number of synthetic images is large or more significant than the number of original real images. This may be because the synthetic images do not completely guarantee the original images' location and quantity of labels. Additionally, as the images produced by the Stable Diffusion \cite{SDXL} model are trained on the LION-5B dataset \cite{lion}, the resulting images do not share the same distribution as the target dataset. So, the synthetic data cannot completely replace the original training dataset used to train the model.

\subsection{Conclusion}
In this study, we introduced a novel data augmentation pipeline for semantic segmentation tasks based on Controllable Diffusion models. Our proposed methods, including \textit{Class-Prompt Appending}, \textit{Visual Prior Blending}, and a \textit{class-balancing algorithm}, effectively address challenges associated with generating synthetic images while preserving the structure and class balance of labeled datasets. By combining synthetic and real-world data, we demonstrated improvements in segmentation performance on the PASCAL VOC datasets in terms of mIoU, compared to training on original datasets alone. These results validate the effectiveness of our approach, particularly in scenarios with limited data availability. Furthermore, our method can be seamlessly combined with other augmentation methods to further enhance performance. Through extensive experiments, we have demonstrated the versatility and robustness of our approach, providing a strong foundation for future research in data augmentation.

\section{\uppercase{Acknowledgments}}

This research is funded by University of Information Technology-Vietnam National University of Ho Chi Minh city under grant number D1-2024-76.

% \section*{\uppercase{Acknowledgements}}

% If any, should be placed before the references section
% without numbering. To do so please use the following command:
% \textit{$\backslash$section*\{ACKNOWLEDGEMENTS\}}

\bibliographystyle{apalike}
{\small
\bibliography{ref}}

\begin{thebibliography}{}

\bibitem[Azizi et~al., 2023]{aug_cls2}
Azizi, S., Kornblith, S., Saharia, C., Norouzi, M., and Fleet, D.~J. (2023).
\newblock Synthetic data from diffusion models improves imagenet classification.
\newblock {\em Transactions on Machine Learning Research}.

\bibitem[Canny, 1986]{canny}
Canny, J. (1986).
\newblock A computational approach to edge detection.
\newblock {\em IEEE Transactions on Pattern Analysis and Machine Intelligence}, PAMI-8(6):679--698.

\bibitem[Chae et~al., 2023]{seggen1}
Chae, J., Cho, H., Go, S., Choi, K., and Uh, Y. (2023).
\newblock Semantic image synthesis with unconditional generator.
\newblock In {\em Thirty-seventh Conference on Neural Information Processing Systems}.

\bibitem[Chan et~al., 2022]{lineart}
Chan, C., Durand, F., and Isola, P. (2022).
\newblock Learning to generate line drawings that convey geometry and semantics.
\newblock In {\em 2022 IEEE/CVF Conference on Computer Vision and Pattern Recognition (CVPR)}, pages 7905--7915.

\bibitem[Chen et~al., 2018]{deeplabv3plus}
Chen, L.-C., Zhu, Y., Papandreou, G., Schroff, F., and Adam, H. (2018).
\newblock Encoder-decoder with atrous separable convolution for semantic image segmentation.
\newblock In Ferrari, V., Hebert, M., Sminchisescu, C., and Weiss, Y., editors, {\em Computer Vision -- ECCV 2018}, pages 833--851, Cham.

\bibitem[Cheng et~al., 2022]{mask2former}
Cheng, B., Misra, I., Schwing, A.~G., Kirillov, A., and Girdhar, R. (2022).
\newblock Masked-attention mask transformer for universal image segmentation.
\newblock In {\em 2022 IEEE/CVF Conference on Computer Vision and Pattern Recognition (CVPR)}.

\bibitem[Everingham et~al., 2015]{voc}
Everingham, M., Eslami, S. M.~A., Gool, L.~V., Williams, C. K.~I., Winn, J., and Zisserman, A. (2015).
\newblock The pascal visual object classes challenge: A retrospective.
\newblock {\em International Journal of Computer Vision}, 111(1):98--136.

\bibitem[Fang et~al., 2024]{aug4}
Fang, H., Han, B., Zhang, S., Zhou, S., Hu, C., and Ye, W.-M. (2024).
\newblock Data augmentation for object detection via controllable diffusion models.
\newblock In {\em 2024 IEEE/CVF Winter Conference on Applications of Computer Vision (WACV)}, pages 1246--1255.

\bibitem[He et~al., 2023]{aug_cls3}
He, R., Sun, S., Yu, X., Xue, C., Zhang, W., Torr, P., Bai, S., and QI, X. (2023).
\newblock {IS} {SYNTHETIC} {DATA} {FROM} {GENERATIVE} {MODELS} {READY} {FOR} {IMAGE} {RECOGNITION}?
\newblock In {\em The Eleventh International Conference on Learning Representations}.

\bibitem[Ho et~al., 2020]{ddpm}
Ho, J., Jain, A., and Abbeel, P. (2020).
\newblock Denoising diffusion probabilistic models.
\newblock In {\em Advances in Neural Information Processing Systems}, volume~33, pages 6840--6851.

\bibitem[Li et~al., 2023]{blip2}
Li, J., Li, D., Savarese, S., and Hoi, S. (2023).
\newblock Blip-2: bootstrapping language-image pre-training with frozen image encoders and large language models.
\newblock In {\em Proceedings of the 40th International Conference on Machine Learning}, ICML'23.

\bibitem[Lin et~al., 2014]{coco}
Lin, T.-Y., Maire, M., Belongie, S., Hays, J., Perona, P., Ramanan, D., Doll{\'a}r, P., and Zitnick, C.~L. (2014).
\newblock Microsoft coco: Common objects in context.
\newblock In {\em Computer Vision -- ECCV 2014}, pages 740--755, Cham. Springer International Publishing.

\bibitem[{MMSegmentation Contributors}, 2020]{mmseg}
{MMSegmentation Contributors} (2020).
\newblock {OpenMMLab Semantic Segmentation Toolbox and Benchmark}.

\bibitem[Mou et~al., 2023]{t2iadapter}
Mou, C., Wang, X., Xie, L., Wu, Y., Zhang, J., Qi, Z., Shan, Y., and Qie, X. (2023).
\newblock T2i-adapter: Learning adapters to dig out more controllable ability for text-to-image diffusion models.

\bibitem[Nguyen et~al., 2023]{datasetsd}
Nguyen, Q.~H., Vu, T.~T., Tran, A.~T., and Nguyen, K. (2023).
\newblock Dataset diffusion: Diffusion-based synthetic data generation for pixel-level semantic segmentation.
\newblock In {\em Thirty-seventh Conference on Neural Information Processing Systems}.

\bibitem[Paszke et~al., 2019]{pytorch}
Paszke, A., Gross, S., Massa, F., Lerer, A., Bradbury, J., Chanan, G., Killeen, T., Lin, Z., Gimelshein, N., Antiga, L., Desmaison, A., Kopf, A., Yang, E., DeVito, Z., Raison, M., Tejani, A., Chilamkurthy, S., Steiner, B., Fang, L., Bai, J., and Chintala, S. (2019).
\newblock Pytorch: An imperative style, high-performance deep learning library.
\newblock In {\em Advances in Neural Information Processing Systems}, volume~32.

\bibitem[Podell et~al., 2024]{SDXL}
Podell, D., English, Z., Lacey, K., Blattmann, A., Dockhorn, T., M{\"u}ller, J., Penna, J., and Rombach, R. (2024).
\newblock Sdxl: Improving latent diffusion models for high-resolution image synthesis.
\newblock In {\em The Twelfth International Conference on Learning Representations}.

\bibitem[Ranftl et~al., 2022]{depth}
Ranftl, R., Lasinger, K., Hafner, D., Schindler, K., and Koltun, V. (2022).
\newblock Towards robust monocular depth estimation: Mixing datasets for zero-shot cross-dataset transfer.
\newblock {\em IEEE Transactions on Pattern Analysis and Machine Intelligence}, 44(3).

\bibitem[Rombach et~al., 2022]{SD}
Rombach, R., Blattmann, A., Lorenz, D., Esser, P., and Ommer, B. (2022).
\newblock High-resolution image synthesis with latent diffusion models.
\newblock In {\em Proceedings of the IEEE/CVF Conference on Computer Vision and Pattern Recognition (CVPR)}, pages 10684--10695.

\bibitem[Schuhmann et~al., 2022]{lion}
Schuhmann, C., Beaumont, R., Vencu, R., Gordon, C.~W., Wightman, R., Cherti, M., Coombes, T., Katta, A., Mullis, C., Wortsman, M., Schramowski, P., Kundurthy, S.~R., Crowson, K., Schmidt, L., Kaczmarczyk, R., and Jitsev, J. (2022).
\newblock {LAION}-5b: An open large-scale dataset for training next generation image-text models.
\newblock In {\em Thirty-sixth Conference on Neural Information Processing Systems Datasets and Benchmarks Track}.

\bibitem[Su et~al., 2021]{sketch}
Su, Z., Liu, W., Yu, Z., Hu, D., Liao, Q., Tian, Q., Pietikäinen, M., and Liu, L. (2021).
\newblock Pixel difference networks for efficient edge detection.
\newblock In {\em 2021 IEEE/CVF International Conference on Computer Vision (ICCV)}, pages 5097--5107.

\bibitem[Trabucco et~al., 2024]{aug_cls1}
Trabucco, B., Doherty, K., Gurinas, M.~A., and Salakhutdinov, R. (2024).
\newblock Effective data augmentation with diffusion models.
\newblock In {\em The Twelfth International Conference on Learning Representations}.

\bibitem[Wang et~al., 2022]{u2pl}
Wang, Y., Wang, H., Shen, Y., Fei, J., Li, W., Jin, G., Wu, L., Zhao, R., and Le, X. (2022).
\newblock Semi-supervised semantic segmentation using unreliable pseudo labels.
\newblock In {\em Proceedings of the IEEE/CVF International Conference on Computer Vision and Pattern Recognition (CVPR)}.

\bibitem[Wu et~al., 2024]{aug_wsss}
Wu, W., Dai, T., Huang, X., Ma, F., and Xiao, J. (2024).
\newblock Gpt-prompt controlled diffusion for weakly-supervised semantic segmentation.

\bibitem[Xie and Tu, 2015]{hed}
Xie, S. and Tu, Z. (2015).
\newblock Holistically-nested edge detection.
\newblock In {\em 2015 IEEE International Conference on Computer Vision (ICCV)}, pages 1395--1403.

\bibitem[Yang et~al., 2023]{inpaint}
Yang, B., Gu, S., Zhang, B., Zhang, T., Chen, X., Sun, X., Chen, D., and Wen, F. (2023).
\newblock Paint by example: Exemplar-based image editing with diffusion models.
\newblock In {\em 2023 IEEE/CVF Conference on Computer Vision and Pattern Recognition (CVPR)}, pages 18381--18391.

\bibitem[Yu et~al., 2020]{bdd}
Yu, F., Chen, H., Wang, X., Xian, W., Chen, Y., Liu, F., Madhavan, V., and Darrell, T. (2020).
\newblock Bdd100k: A diverse driving dataset for heterogeneous multitask learning.
\newblock In {\em IEEE/CVF Conference on Computer Vision and Pattern Recognition (CVPR)}.

\bibitem[Zhang et~al., 2023]{controlnet}
Zhang, L., Rao, A., and Agrawala, M. (2023).
\newblock Adding conditional control to text-to-image diffusion models.
\newblock In {\em 2023 IEEE/CVF International Conference on Computer Vision (ICCV)}, pages 3813--3824.

\bibitem[Zhao et~al., 2017]{pspnet}
Zhao, H., Shi, J., Qi, X., Wang, X., and Jia, J. (2017).
\newblock Pyramid scene parsing network.
\newblock In {\em Proceedings of the IEEE Conference on Computer Vision and Pattern Recognition (CVPR)}.

\bibitem[Zhou et~al., 2019]{ade20k}
Zhou, B., Zhao, H., Puig, X., Xiao, T., Fidler, S., Barriuso, A., and Torralba, A. (2019).
\newblock Semantic understanding of scenes through the ade20k dataset.
\newblock {\em International Journal of Computer Vision}, 127(3):302--321.

\end{thebibliography}

% \bibliographystyle{apalike}
% {\small
% \bibliography{example}}

% \section*{\uppercase{Appendix}}

% If any, the appendix should appear directly after the
% references without numbering, and not on a new page. To do so please use the following command:
% \textit{$\backslash$section*\{APPENDIX\}}

\end{document}